\documentclass[sigconf]{acmart}
\usepackage[utf8]{inputenc}
\usepackage{bm}
\usepackage{array}
\usepackage{float}
\usepackage{natbib}

\AtBeginDocument{%
  \providecommand\BibTeX{{%
    \normalfont B\kern-0.5em{\scshape i\kern-0.25em b}\kern-0.8em\TeX}}}

\copyrightyear{2020}
\acmYear{2020}
\setcopyright{acmcopyright}\acmConference[SBES '20]{34th Brazilian Symposium on Software Engineering}{October 21--23, 2020}{Natal, Brazil}
\acmBooktitle{34th Brazilian Symposium on Software Engineering (SBES '20), October 21--23, 2020, Natal, Brazil}
\acmPrice{15.00}
\acmDOI{10.1145/3422392.3422462}
\acmISBN{978-1-4503-8753-8/20/09}

\begin{document}

\title{CoNCRA: A Convolutional Neural Network Code Retrieval Approach}

\author{Marcelo de Rezende Martins}
\email{rezende.martins@gmail.com}
\affiliation{%
\institution{IPT - Institute for Technological Research}
 \city{Sao Paulo}
 \state{Sao Paulo}
 \country{Brazil}
}

\author{Marco Aurélio Gerosa}
\email{marco.gerosa@nau.edu}
\affiliation{%
 \institution{Northern Arizona University (NAU)}
 \city{Flagstaff}
 \state{Arizona}
 \country{United States}
}


\begin{abstract}
Software developers routinely search for code using general-purpose search engines. However, these search engines cannot find code semantically unless it has an accompanying description. We propose a technique for semantic code search: A Convolutional Neural Network approach to code retrieval (CoNCRA). Our technique aims to find the code snippet that most closely matches the developer's intent, expressed in natural language. We evaluated our approach's efficacy on a dataset composed of questions and code snippets collected from Stack Overflow. Our preliminary results showed that our technique, which prioritizes local interactions (words nearby), improved the state-of-the-art (SOTA) by 5\% on average, retrieving the most relevant code snippets in the top 3 (three) positions by almost 80\% of the time. Therefore, our technique is promising and can improve the efficacy of semantic code retrieval.
\end{abstract}



\maketitle

\section{Introduction}

The advent of open-source software and question and answering websites contributed for improving the way developers produce code. Nowadays, code search permeates the development activities. Developers can spend 15\% of their time searching online for how a piece of code works, how to fix a bug, and how to use an API \cite{what-developers-search-for-on-the-web:xia:2017}. According to \citet{sadowski-how-developers-search-for-code-case-study:2015}, at Google, developers search for code 12 times a day, clicking on 2 to 3 results in average per search session.  

Most developers use general-purpose search engines (GPSE) to look for code (e.g., Google Search), which uses page rank and other indexes tactics that are not optimized for searching code. Then, general-purpose search engines do not adequately find code snippets unless they have accompanying descriptions. According to \citet{masudur-developers-use-google-code-retrieval:2018}, developers spend more time, visit more pages, and change queries more often when they are doing code-related searches. In particular, newcomers to a project can greatly benefit from semantic search since they face a variety of entrance barriers \cite{steinmacher2015social}. 

GitHub, a popular source code hosting platform, has attempted to build a semantic code search. They extracted millions of lines of code from its repositories and matched each code snippet to a docstring. The final results were not satisfactory as the tool could find a relevant code snippet only if the user provided a query that matched the docstring description \citep{husain-github-semantic-search-code-2019}. According to \citet{cambronero-deep-code-search-2019}, users' intents were better matched to questions collected from question-answering sites related to programming, e.g., Stack Overflow. Those sites allow users to ask a question and approve the best answer for it. Other users vote for the most helpful answer and mark the wrong or not helpful ones. Those collective actions curate and organize information.

Initial code search studies were based on deductive-logic rules and manually extracted features \cite{Allamanis:2018:SML}. The recent success of artificial neural networks has shifted recent works to a machine learning-based approach. \citet{cambronero-deep-code-search-2019} coined a name, neural code search, i.e., code search based on neural networks.

Recent works applied neural networks to summarize and retrieve code snippets. \citet{cambronero-deep-code-search-2019} proposed a neural network with attention mechanism and \citet{Gu-deep-code-search:2018} presented a recurrent neural network. Our novel approach is based on Convolutional Neural Networks (CNNs). For the best of our knowledge, CNNs have not yet been used to search for code, but have achieved good results in selecting answers \citep{feng-2015, wen-joint-modeling-question-answer-2019}. CNNs prioritize local interactions (words nearby) and its translation invariant, which are important traits for our task. 

In our study, we answer the following research questions:

\begin{itemize}
    \item What is the efficacy of the CONCRA technique? 
    \item How does it compare to the baseline and SOTA methods?
\end{itemize}

\section{Background}

According to \citet{cambronero-deep-code-search-2019}, the main goal of code retrieval is to retrieve code snippets from a code corpus that most closely match a developer's intent, which is expressed in natural language.

The first studies for code search were based on deductive-logic rules and manually extracted features. The Deductive-logic approach, e.g., boolean model, finds a code that precisely matches the keywords expressed in the query. According to \citet{yan-benchmark-code-search-information-retrieval-deep-learning:2020}, these approaches are good at finding API calls and error messages, but struggle to find reusable code and examples that do not have an exact match between the code and query. 

\begin{figure}[h]
  \includegraphics[width=0.45\textwidth]{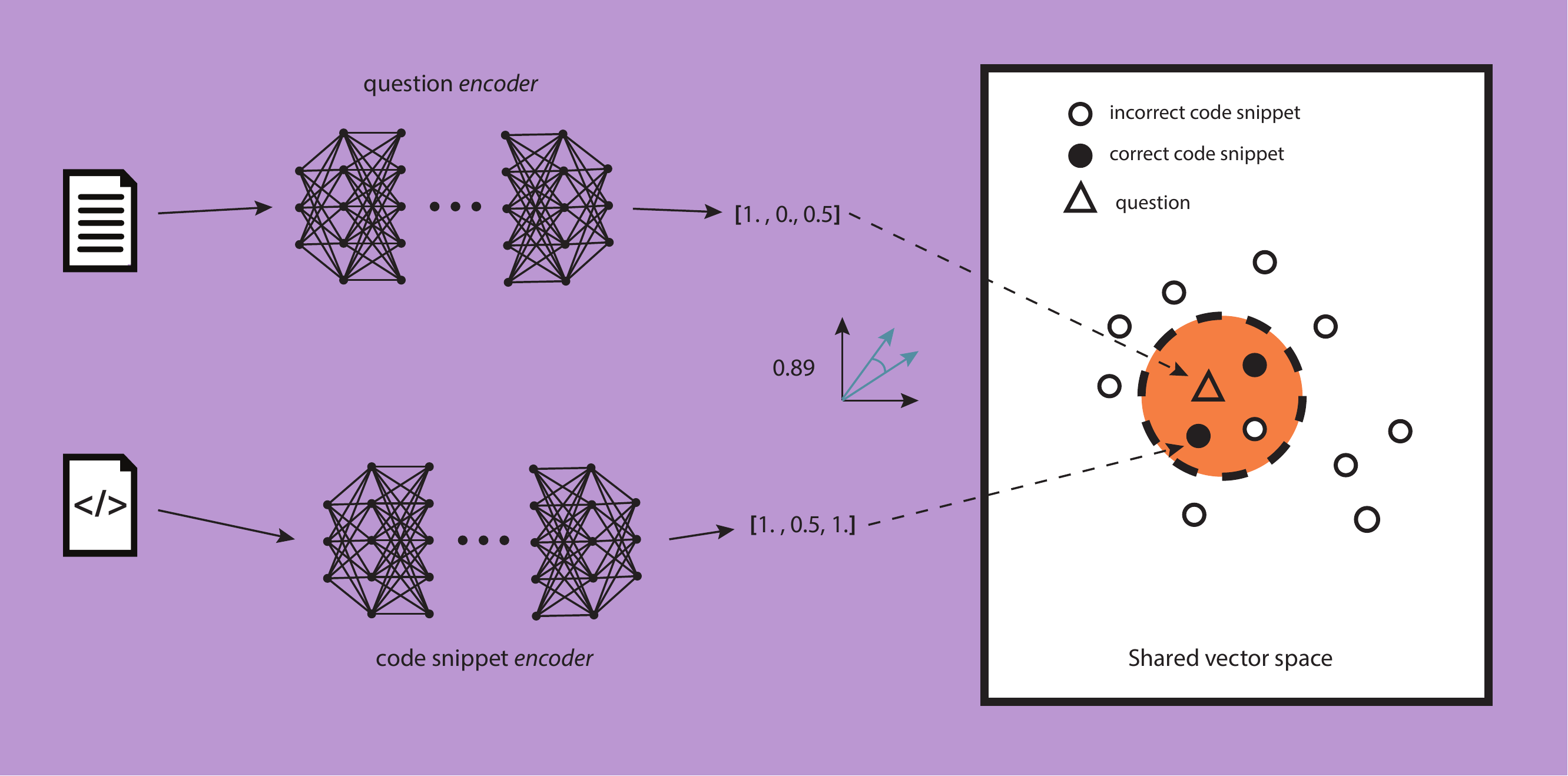}
  \caption{Illustration of the \emph{joint embedding} technique for code retrieval. Two neural networks map a question and a code snippet into a common vector space. The distance between the vectors reflects the relevance of a code snippet to a question. Adapted from \cite{cambronero-deep-code-search-2019}.}
  \Description{None}
  \label{fig:joint-embedding}
\end{figure}

Since neural networks showed good results at translation, question-answering, and classifications tasks in natural language processing, recent works adopted the neural networks for code search. These studies aim to discriminate relevant code snippets from non-relevant ones based on the user's intent. In pursuance of that, code retrieval is reduced to a ranking problem, in which neural networks aim to place code snippets that closely match the developer's intent in the top positions. The most common strategy is \emph{joint embedding}. Joint embedding maps heterogeneous data into a common vector space, in which the distance between embedded input reflects the similarity between the underlying items \cite{li-joint-embedding-images-2015} (see Figure~\ref{fig:joint-embedding}).

To apply joint embedding, one needs to consider word, sentence, and joint embedding. Embedding refers to a continuous vector in a lower dimensional vector space. A function that maps an input to a continuous vector is called \emph{encoder}. Given an input set $X$, an encoder function $F$ can be defined as \cite{cambronero-deep-code-search-2019}:

\begin{equation}
    F: X \to E
\end{equation}

For code retrieval, $X$ can be a set of questions or code snippets and $E$ is a set of continuous vectors or embeddings, such that $E \subset R^{d}$, where $d$ is the dimension. The main goal is to learn two encoders $F$ and $G$ that map a question and a code snippet, respectively, into a common vector space, so that the distance between the vectors reflects the relevance of a code snippet to a question.

\section{Methodology}
\label{sec:methodology}

In this work, we propose the use of convolutional neural networks to learn sentence embedding, i.e., convolutional neural networks will encode the question and code snippet into a continuous vector in a shared vector space. In the following, we explain how words are embedded and what objective function we use, as the objective function tells how neural networks should approximate questions and code snippets.

\subsection{Word embedding}
\label{sec:word-embedding}

The words and terms of a question and code snippet must be encoded into a numeric vector. The most common encoder for words is \emph{word2vec}, which embeds a word into a continuous vector based on the distributional hypothesis. The distributional hypothesis says that two words are similar if they appear together frequently in different contexts \citep{Goodfellow-et-al-2016}. Context can be a sentence, paragraph, or document in NLP tasks. In our case, the context is questions and code snippets.

Word2vec has two strategies: continuous-bag-of-words (CBoW) and skip-gram. The main difference is that CBoW predicts a target word given context words, and skip-gram predicts the context words given a single word. According to \citet{mikolov2013distributed}, CBoW showed good results at syntactic tasks, e.g., finding a superlative of a word or identifying an adverb. At the same time, skip-gram presented good performance for semantic tasks, e.g., finding a state's capital or grouping feminine and masculine words. 

\begin{figure}[h]
\includegraphics[width=0.45\textwidth]{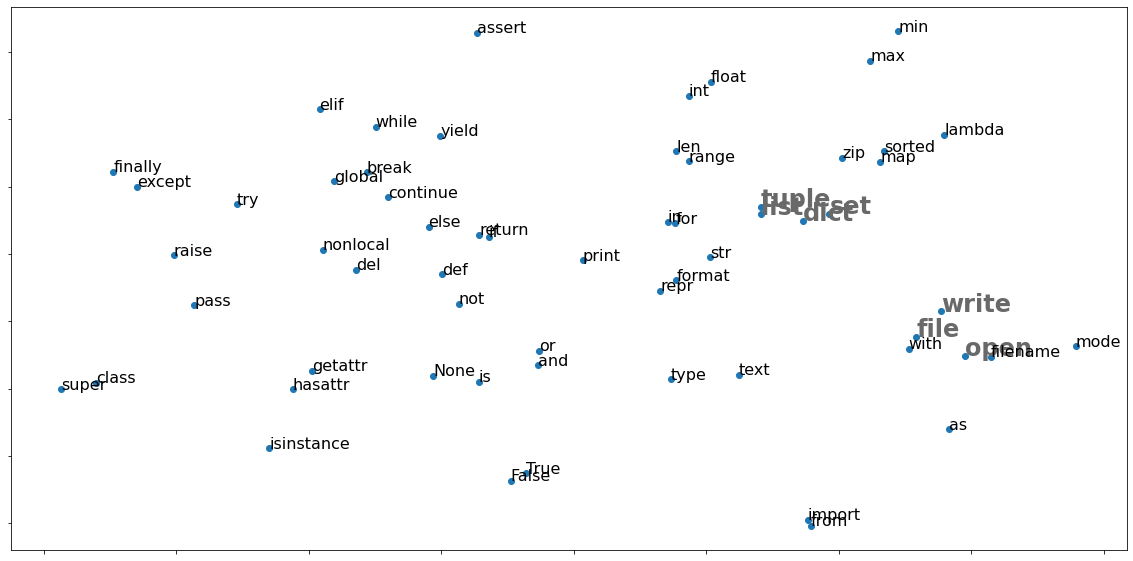}
\caption{2D picture of continuous vectors of the 66 most frequent words from a Python corpus $V$. The illustration was generated by t-SNE, which allows 2D visualization from high-dimensional data. We applied word2vec with skip-gram and a parameter window $5$.}

\label{fig:tsne-code-snippet-python}
\end{figure}

In our work, we opted for skip-gram as a semantic trait is preferable. Semantic trait can help the neural networks to discriminate conditional clauses (e.g., \emph{if}, \emph{elsif}) and loop iteration (e.g., \emph{for}, \emph{while}), for example. Figure~\ref{fig:tsne-code-snippet-python} shows an application of word2vec in a Python related corpus. We can see the similarities between \emph{file}, \emph{write}, and \emph{open}, and \emph{set}, \emph{list}, and \emph{dict} based on the distance between them.

\subsection{Sentence embedding}

We can combine the word embeddings to obtain a sentence embedding. We combine word embeddings by using convolutional neural networks. Convolutional neural networks showed good results at answer selection tasks in NLP---given a question and a set of answers, the model ranks the best answers. Convolutional networks prioritize local interactions (e.g., words nearby) and cannot capture long-range dependencies (e.g., distant words in a sentence). However, this issue is mitigated for code retrieval since most questions and code snippets are short in length.

Given an sentence $\bm{x} = \{ \bm{x}(0), \bm{x}(1), . . ., \bm{x}(n - 1) \}$, such that $\bm{x}(i) \in \mathbb{R}^{d}$ is a continuous vector that represents the $i^{th}$ word of the sentence. Convolutional neural networks combine the elements of vector $\bm{x}$ by applying 2 basic operations: Convolution operation and Maxpool.

A convolution operation uses a filter $\bm{F}  = [\bm{F}(0),· · ·, \bm{F}(m - 1)]$, such that $\bm{F} \in \mathbb{R}^{m X d}$. The operation applies the filter in $m$ words (window size) to produce a new vector. Suppose $\bm{x}(i, i + j)$ refers to a concatenation of the vectors $\bm{x}(i), \bm{x}(i + 1), . . ., \bm{x}(i + j)$. If we apply $\bm{F}$ to $\bm{x}(i, i + m - 1)$, then we can calculate a new vector $\bm{c}(i)$ by:

\begin{equation}\label{eq:calc_convolution_ci}
    \bm{c}(i) = tanh \left[\left(\sum_{j=0}^{m - 1} \bm{x}(i + j)^{T}\bm{F}(j)\right) + b\right]
\end{equation}

In the equation~\ref{eq:calc_convolution_ci}, $\bm{F}$ and $b$ are learnable weights and bias, respectively. The convolution operation slides the filter $\bm{F}$ across the height of input $\bm{x}$ and computes the dot product between the entries of the filter and the input \cite{karpathy-course-cnn-2016}. The operation returns a feature map $\bm{c}$.

\begin{equation}
    \bm{c} = \{ \bm{c}(0), \bm{c}(1), . . ., \bm{c}(n - m) \} 
\end{equation}

The feature map (or activation map) contains the latent and most important features of a sentence. A convolutional neural network may contain thousands of filters, each one extracting specific \emph{m-gram} features (e.g., a filter of $m$ size 2 extracts bigram features). The quantity of feature maps is $|F|$, i.e., the number of filters. After the convolution operation, a pooling layer operates independently on every feature map resizing it spatially, using a max operation \cite{karpathy-course-cnn-2016}. The max operation is applied along the $axis=0$ to produce the final vector $\bm{o}$:

\begin{equation}
    \bm{o} = max\left(\left[\bm{c}_{1}, \bm{c}_{2}, . . ., \bm{c}_{|F|}\right], axis = 0\right)
\end{equation}

\begin{figure}[h]
    \centering
    \includegraphics[width=0.45\textwidth]{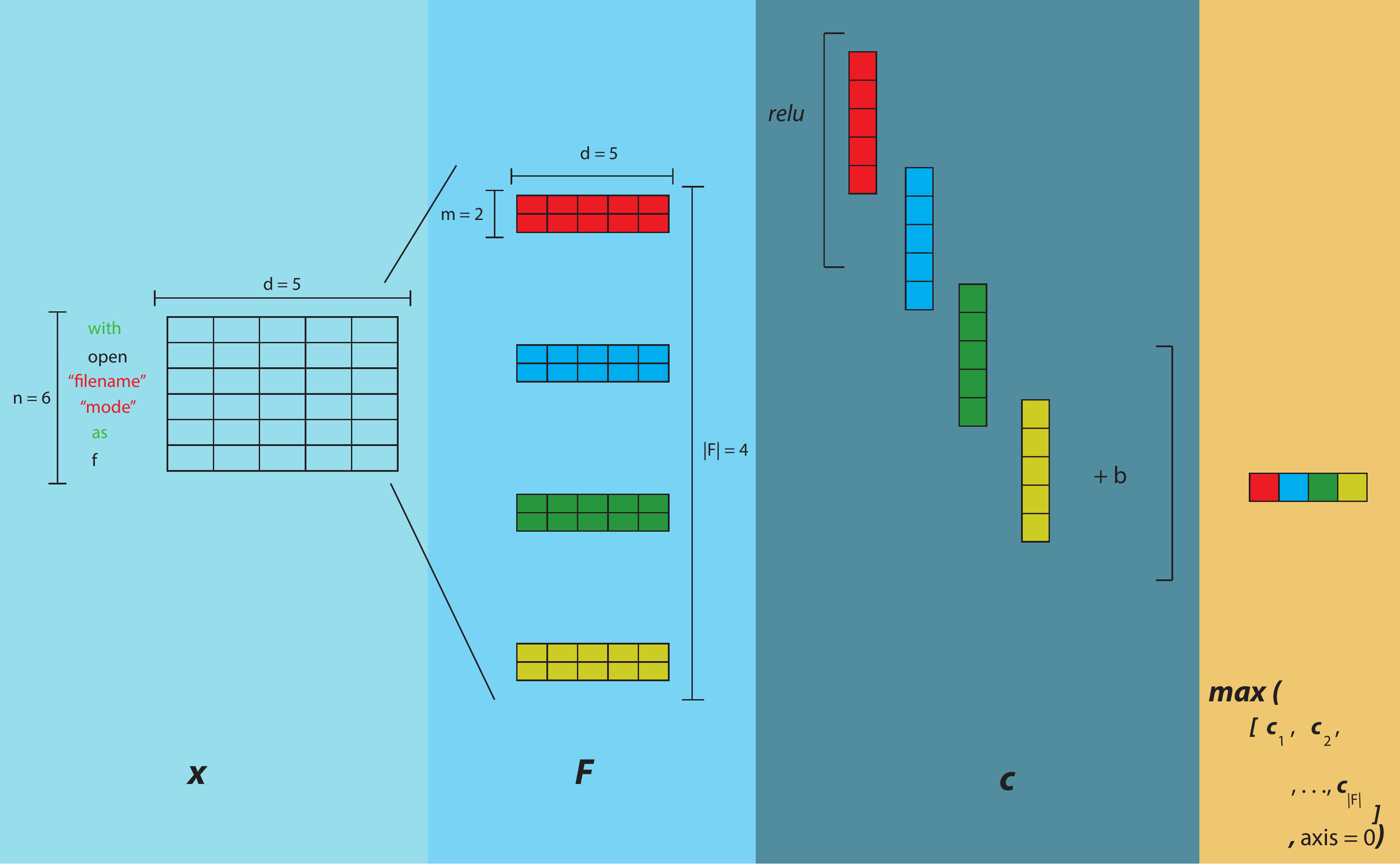}
    \caption{Schematic drawing of our Convolutional Neural Networks (CNN) operations. Our example shows 4 filters $\bm{F} \in \mathbb{R}^{m X d X f}$ with window size $m = 2$. We slide each filter across the height of the input $\bm{x} \in \mathbb{R}^{n X d}$, where $n = 6$ and $d = 5$, and computes the dot product between the entries of the filter and the input \cite{karpathy-course-cnn-2016}. It returns a feature map $\bm{c}$. In the end, a max pooling layer resizes every feature map spatially, obtaining the final vector $\bm{o}$. We use the vector $\bm{o}$ as our question and code snippet embedding. Adapted from \cite{zhang-guide-convolutional-cnn-embedding-ilustration:2015}.}
    \label{fig:cnn-steps-word-embedding}
\end{figure}

The max-pooling helps the convolutional neural networks to be translation invariant. Regardless of the word's position shift, the max-pooling selects the most relevant features and inserts it into the final vector. Figure~\ref{fig:cnn-steps-word-embedding} shows the results of each operation. We use the vector $\bm{o}$ as our sentence embedding, then $\bm{o}$ represents a question or a code snippet, in our case. 

\subsection{Joint embedding}
\label{sec:joint-embedding}

We treat code retrieval as a ranking problem, in which a model should rank relevant code in the top positions based on the developer's intent. To do so, we adopted an objective function that prioritizes the relative preference of the code snippets. In our case, the objective function helps the neural networks to separate the correct answers from incorrect ones during the training phase.

Given a question and code snippet set, $\mathbb{Q}$ and $\mathbb{C}$, our training input comprises a triple $<\bm{q}, \bm{c^{+}}, \bm{c^{-}}>$, where $\bm{c^{+}} \in \mathbb{C}$ indicates a correct code snippet for a question $\bm{q} \in \mathbb{Q}$ and $\bm{c^{-}} \in \mathbb{C}$ an incorrect one sampled from the training data. We used the hinge loss as our objective function. Formally, for a triple $<\bm{q}, \bm{c^{+}}, \bm{c^{-}}>$, the definition of hinge loss is:

\begin{equation}
J = max(0, m - h_{\theta}(\bm{q}, \bm{c^{+}}) + h_{\theta}(\bm{q}, \bm{c^{-}}))
\end{equation}

$m$ is a margin and $h_{\theta}$ is a similarity function (e.g., \textit{cosine}). During the training phase, the goal is to minimize the cost function $J$. To obtain that, the model aims to satisfy the following condition: $h_{\theta}(\bm{q}, \bm{c^{+}}) - h_{\theta}(\bm{q}, \bm{c^{-}}) \geq m$. Then, the hinge loss function induces our model to score $c^{+}$ higher than $c^{-}$ for a given margin $m$. For the similarity function $h_{\theta}$, we used \emph{cosine}.

\section{Experiments}

To verify the efficacy of our approach and how it compares to other approaches, we trained and evaluated all models in the same environment, following \citet{yao-2018} and \citet{ iyer-etal-2016-summarizing} experiment protocol. We compared the following approaches:

\begin{itemize}
    \item \emph{CoNCRA}: our proposed approach, described in Section~\ref{sec:methodology}. We tried two variations of our approach, the Convolutional Neural Networks (CNN) and Shared CNN. Their difference is that Shared CNN shares the weights by learning the question and code snippet embeddings, while CNN learns different weights for each one.
    \item \emph{Embedding}: it is our baseline model. It is a simple architecture that applies a max-pooling layer to the word embeddings. 
    \item \emph{Unif}: it is the solution proposed by \citet{cambronero-deep-code-search-2019}. They used two distinct layers for the question and code snippet embedding. They applied an average pooling layer to word embedding in order to learn the question embeddings. For the code snippet, they used an attention mechanism, which applies a weighted average to each word embedding, ''giving attention'' to the most relevant word of the code.
\end{itemize}

We evaluated the models on the StaQC dataset, a systematically mined question-code dataset from Stack Overflow \cite{yao-2018}. The main difference of StaQC to other datasets is that it is composed of ''how-to-do-it'' questions, as most of the answers to those types of questions are straightforward. StaQC contains SQL and Python questions, but, in our experiments, we used Python questions only. 

\begin{table}[h]
\centering
\begin{tabular}{ p{5cm} c c }
\hline
  & \multicolumn{2}{c}{\textbf{Question}}\\
\hline
\textbf{Code snippet} & \textbf{Python} & \textbf{SQL}  \\
\hline

$N_{1}$: Single code snippet in the answer description & $85.294$ & $75.637$ \\

$N_{2}$: Automatically annotated code snippets & $60.083$ & $41.826$ \\

$N_{3}$: Manually annotated code snippets & $2.169$ & $2.056$  \\

 \hline
 \textbf{Total} & $\bm{147.546}$ & $\bm{119.519}$\\
 \hline 

\end{tabular}
\caption{Summary of StaQC dataset \cite{yao-2018}. Questions from $N2$ sample ("Automatically annotated code snippets") may contain more than one code snippet per answer description. Some code snippets may not be a solution to the question. So, the authors proposed a framework to annotate the code snippets automatically and it could achieve an F1 score of $0,916$ and an accuracy of $0,911$.}
\label{table:summary-training-data-yao-staqc}
\end{table}

\begin{table}[h]
\centering
\begin{tabular}{ l r  }
 \hline
 \textbf{Sample} & \textbf{Quantity of pairs $<q_{i}, c_{i}^{+}>$}\\
 \hline
 $N_{2} = \text{Training}$ & $60.083$\\
 
 $N_{3} \supset \text{DEV}$ & $1.085$ \\
 
 $N_{3} \supset \text{EVAL}$ & $1.084$\\
 \hline
 \textbf{Total} & $\bm{62.252}$\\
 \hline
\end{tabular}
\caption{Summary of our training and evaluation samples. The samples are composed of a pair $<q_{i}, c_{i}^{+}>$, where $q_{i}$ is a question and $c_{i}^{+}$ is a code snippet annotated as solution. We split the manually annotated dataset into two parts: DEV and EVAL, according to \cite{iyer-etal-2016-summarizing} procedure. }
\label{table:training-sample-division}
\end{table}

All models were trained on sample $N2$ (see Table~\ref{table:summary-training-data-yao-staqc} and Table~\ref{table:training-sample-division}), because 27\% of the questions contain more than one annotated answer, leading to more variance in our training dataset. The training and evaluation follow \citet{iyer-etal-2016-summarizing}'s procedure, in which the model is evaluated on a manually annotated dataset each epoch according to a Mean Reciprocal rank (MRR). The MRR tells if a model ranked the annotated answer in higher positions, i.e., higher values for MRR indicate that the accepted answers were ranked in the top positions.

For the training phase, we split 70\% of $N2$ sample for training and 30\% for validation. We run the models for 500 epochs and stop early if the training loss ($J$) is less than $0.0001$ or the validation loss does not improve after 25 consecutive epochs. For the final evaluation, we choose the best model of the training phase according to MRR and run it for 20 iterations in the $N3$ sample (see Table~\ref{table:training-sample-division}). The final result is the average of the MRR for each pair $<q_{i}, c_{i}^{+}>$ of $N3$ and other 49 distractors $c_{j}$, which were selected randomly from the training sample, such that $c_{i}^{+} \neq c_{j}$.

\begin{figure}[h]
  \includegraphics[width=0.45\textwidth]{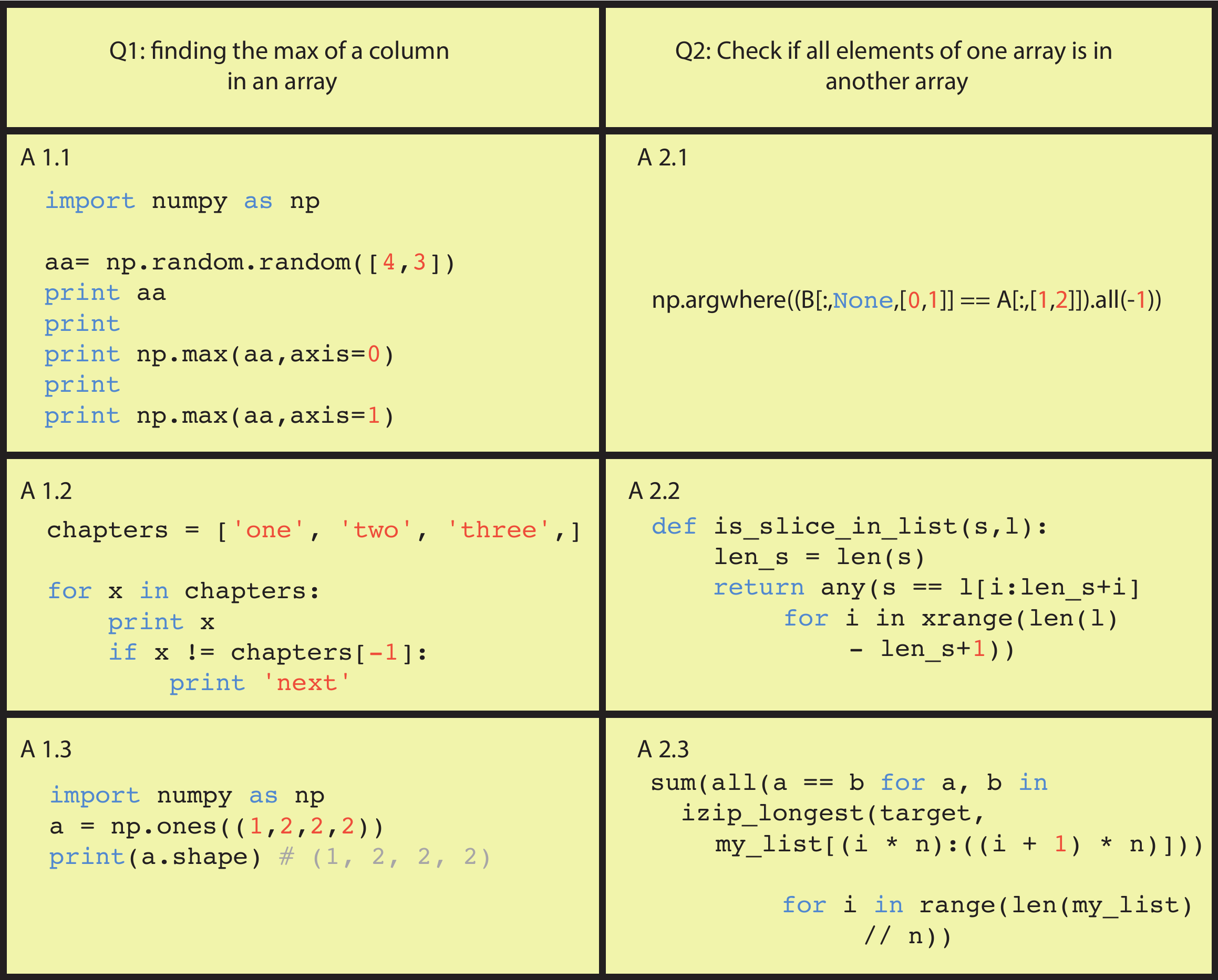}
  \caption{Example of questions and  that our model (CoNCRA) gave. Our model answered the first question (Q1) correctly, selecting an answer (A 1.1) based on the Numpy library, which adds support for multi-dimensional and large arrays in Python. The second answer (A 1.2) for the first question returns the last element of an array, which is incorrect, but seems interesting. The third one (A 1.3) shows the array's dimension using the Numpy library, so our model found a correlation between Numpy and array operations. In the other example (Q2), the first answer is incorrect, as it checks the presence of elements in a matrix, not an array. Again, our model showed a Numpy answer. The second one (A 2.2) is the correct answer, and the third one (A 2.3) checks how many times a sequence repeats in a data frame. }
  \Description{None}
  \label{fig:concrete-examples}
\end{figure}

We provide the source code for our preliminary experiments in the following repository: \url{https://github.com/mrezende/concra}. The repository contains our proposed model, the baseline ones, training, and evaluation source code. We also provide the original and pre-processed dataset. The source code is written in Python, version 3.6.9, and we used the libraries Keras (version 2.2.4-tf) and TensorFlow (1.15.2). The tests were all conducted in the Colab platform\footnote{\url{https://colab.research.google.com/}}. In Figure \ref{fig:concrete-examples}, we provide examples of the output of our tool.

\subsection{Results}

Table~\ref{table:resultados} compiles the final results after 20 runs on the $N3$ sample. Shared CNN with 4000 filters got the best results (row D3 and F3). Our proposed architecture achieved an MRR score 5\% higher on average than the best result obtained by Unif (row B1), which can be considered a state-of-the-art approach. Our MRR result is 11\% higher than the baseline model (row A1). 

\begin{table*}[t]
\centering
\begin{tabular}{ p{1cm} p{6cm} >{\raggedleft\arraybackslash}p{4cm} >{\raggedleft\arraybackslash}p{4cm} }
 \hline
    & & \multicolumn{2}{c}{\textbf{Results}}\\
 \hline
 & \textbf{Models} & \textbf{MRR} & \textbf{TOP-1}\\
 \hline
 A1 & Embedding (m = $0.1$) & $0.637$& $0.493 \pm 0.009$\\
 
 \hline
 
 B1 & Unif (m = $0.2$) & $0.675 \pm 0.006$ & $0.539 \pm 0.009$\\
 
 \hline
 
 C1 & CNN / F = 1000 & $0.669 \pm 0.006$ & $0.527 \pm 0.012$\\
 
 C2 & CNN / F = 2000 & $0.673 \pm 0.007$ & $0.531 \pm 0.012$\\
 
 C3 & CNN / F = 4000 & $0.687 \pm 0.006$ & $0.553 \pm 0.011$\\
 
 \hline
 
 D1 & Shared CNN / F = 1000 & $0.678 \pm 0.007$ & $0.548 \pm 0.012$\\
 
 D2 & Shared CNN / F = 2000 & $0.694 \pm 0.008$ & $0.565 \pm 0.012$\\
 
 D3 & Shared CNN / F = 4000 & $0.700 \pm 0.004$ & $0.569 \pm 0.009$\\
 
 \hline
 
 E1 & CNN with BN / F = 1000 & $0.682 \pm 0.007$ & $0.543 \pm 0.012$\\
 
 E2 & CNN with BN / F = 2000 & $0.689 \pm 0.006$ & $0.553 \pm 0.011$\\
 
 E3 & CNN with BN / F = 4000 & $0.688 \pm 0.006$ & $0.553 \pm 0.011$\\
 
 \hline
 
 F1 & Shared CNN with BN / F = 1000 & $0.690 \pm 0.008$ & $0.553 \pm 0.015$\\
 
 F2 & Shared CNN with BN / F = 2000 & $0.700 \pm 0.007$ & $0.573 \pm 0.012$\\
 
 F3 & Shared CNN with BN / F = 4000 & $0.701 \pm 0.008$ & $0.577 \pm 0.015$\\
 
\hline
\end{tabular}
\caption{The experimental results of EVAL sample for Embedding, Unif, and our two CNN variations: CNN and Shared CNN. \emph{m} refers to the margin loss of the hinge loss function (lines A1 and B1). \emph{F} indicates the number of filters. BN is an acronym of Batch Normalization. Our CNN architecture used a margin loss $m = 0.05$ and a window size of $2$.}
\label{table:resultados}
\end{table*}

As presented in Table~\ref{table:resultados}, an increase in the number of filters resulted in a better performance for CNN, as the model capacity and the number of extracted features grow (row D3 and F3). We tried different margin loss (see Section~\ref{sec:joint-embedding}). CNN got the best results with $0.05$, while Unif and Embedding got better results with $0.2$ and $0.1$, respectively. 

We verified that shared weights models (row D and F) got better results than independently weights ones (row C and E). One reason is that the optimizer of independent weights architecture has to learn the double of parameters, increasing the learning difficulty \cite{feng-2015}. We also tried a batch normalization technique to avoid overfitting and help our models learn quickly and a more stable way, but only CNN and Shared CNN got better results. For Unif and Embedding, it did not improve the performance at all. 

\begin{figure}[h]
    \centering
    \includegraphics[width=0.45\textwidth]{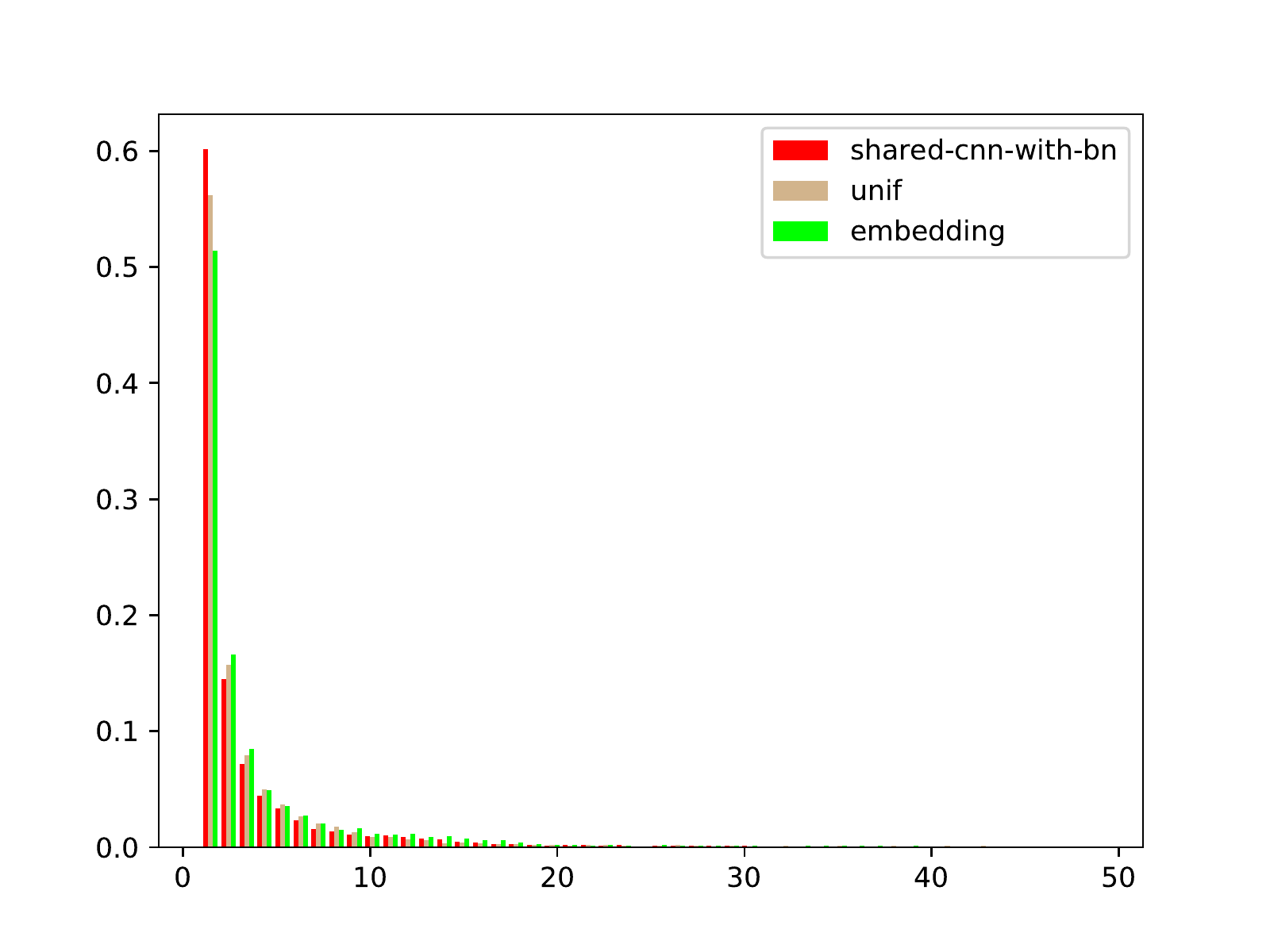}
    \caption{Histogram of the first positions observed for the annotated code snippet during the final evaluation. The labels \emph{shared-cnn-with-bn}, \emph{unif}, and \emph{embedding} refer to lines F3, A1, and B1, respectively, in Table~\ref{table:resultados}.}
    \label{fig:histogram-mrr}
\end{figure}

Figure~\ref{fig:histogram-mrr} illustrates the MRR (Mean Reciprocal Rank), showing the first position of the annotated code snippet during the final evaluation.

Both CNN and Unif ranked the code snippets among the first three positions in 75\% of cases. We got a TOP-1 accuracy of 60\%, i.e., we ranked the relevant code snippet in the first place in 60\% of cases. Unif and Embedding got a TOP-1 accuracy of almost 50\%. However, it is worth noticing that MRR considers only the annotated code snippet position. If the model shows up another code snippet, which correctly answers the question, our metric does not consider it and the model is penalized. 

\subsection{Threats to validity}

We trained the models on the StaQC dataset, a systematically mined Stack Overflow dataset. The authors used neural networks to annotate the code snippets trained on a manual dataset. In our case, we trained the models on the automatically annotated corpus and evaluated on the manual ones. To mitigate bias, we adopted \citet{iyer-etal-2016-summarizing}'s procedure for training and evaluation.

\section{Related Work}

We summarize the difference between our work and related work in Table~\ref{table:summary-joint-embedding-related-work}. Most works differ on how they combine the word embeddings to obtain a sentence embedding. Recent works (row E and F) used Skip-gram (see Section~\ref{sec:word-embedding}) and adopted simpler architectures for sentence embedding than previous work (row C and D).  

\begin{table}[t]
\centering
\begin{tabular}{ p{0.1cm} p{2cm} p{1.5cm} p{1.5cm} p{1.5cm} }
 \hline
 & \textbf{Work} & \textbf{Feature} & \textbf{Word Embedding} & \textbf{Sentence Embedding} \\
 \hline
A & \citet{Allamanis-bimodal-source-code-natural-language:2015} & Token / Parse Tree & Probabilistic Model & Average / Context matrix  \\

B &\citet{iyer-etal-2016-summarizing} & Token & One-hot encoding & LSTM with attention mechanism  \\

C &\citet{Chen-bi-variational-autoencoder:2018} & Token & Bi-VAE & Bi-VAE, Average and MLP  \\

D &\citet{Gu-deep-code-search:2018} & Token / Method name, API invocation and Token & bi-LSTM & Max-pooling / Max-pooling and MLP   \\

E &\citet{Sachdev-neural-code-search:2018} & Token & Skip-gram & Average / TF-IDF   \\

F &\citet{cambronero-deep-code-search-2019} & Token & Skip-gram & Average / Attention   \\

G & \textit{Our work} & \textit{Token} & \textit{Skip-gram} & \textit{CNN}   \\

 \hline
\end{tabular}
\caption{Summary of related joint embedding work. The column \emph{Feature} refers to question and code representation. Different approaches for a question and code snippet are separated by slash (''/''), showing the question method first followed by the code snippet technique. Adapted from \cite{yan-benchmark-code-search-information-retrieval-deep-learning:2020}.}
\label{table:summary-joint-embedding-related-work}
\end{table}

Previous work (rows D to F) that used GitHub data extracted the methods from the source code and matched them to docstring descriptions. Works that used Stack Overflow data (rows A to C) paired question titles to the accepted answers' code snippet. For the search corpus, some works (rows D to F) adopted a GitHub corpus with millions of pieces of code, while others (rows A to C) retrieved code snippets from a small sample of 50 randomly selected code snippets. These studies (rows A to G) are commonly evaluated using questions collected from Stack Overflow. 

Although our architecture is more complex, since it requires more parameters and time to train, than \citet{cambronero-deep-code-search-2019}'s (row F) architecture (SOTA), we can train our model offline. We matched question title to code snippets collected from Stack Overflow following \citet{Allamanis-bimodal-source-code-natural-language:2015} and \citet{iyer-etal-2016-summarizing} (rows A and B) work. 

\section{Conclusion}

Our model, CoNCRA, achieved an MRR score 5\% higher on average than Unif, a state-of-the-art technique. We could rank the most relevant code snippet among the first 3 (three) positions in 78\% of cases. Our technique achieved a TOP-1 accuracy of 60\%, while the other techniques achieved 50\%. 

The results seem promising, and we plan further investigation to check if our model is invariant to other datasets. We will also investigate transfer learning, e.g., checking if a model trained on a Stack Overflow dataset (curated corpus) can find relevant code snippets in a GitHub corpus (non-curated one). Our approach is based on an NLP technique proposed by \citet{feng-2015}, and future work may use transformers and autoencoders, as those techniques showed good results in many NLP tasks. 
\section{Acknowledgement}
This work was partially supported by the National Science Foundation (grant 1815503).

\bibliographystyle{ACM-Reference-Format}
\bibliography{sample-base}


\end{document}